\documentclass{article}

\title{The Logovista English--Japanese Machine Translation System}
\author{Barton D. Wright}
\date{2026}

\begin{document}
\maketitle

\begin{abstract}
This paper documents the architecture, development practices, and preserved artifacts of the Logovista English--Japanese machine translation system, a large, explicitly rule-based MT system that was developed and sold commercially from the early 1990s through at least 2012. The system combined hand-authored grammatical rules, a large central dictionary encoding syntactic and semantic constraints, and chart-based parsing with weighted interpretation scoring to manage extensive structural ambiguity.

The account emphasizes how the system was extended and maintained under real-world usage pressures, including regression control, ambiguity management, and the limits encountered as coverage expanded. Unlike many rule-based MT systems described primarily in research settings, Logovista was deployed for decades and evolved continuously in response to practical requirements. The paper is intended as a technical and historical record rather than an argument for reviving rule-based MT, and describes the software and linguistic resources that have been preserved for potential future study.
\end{abstract}

\section{Purpose and Scope}

This document is intended as a technical and historical record, not as an argument for reviving rule-based machine translation. The system described here was used and sold commercially for roughly two decades, and the surviving archive includes source code and version-control logs recording changes made under ongoing real-world use. This material may be useful in understanding how a large, explicitly rule-based MT system adapted over time in response to practical pressures, and where such approaches ultimately encountered limits.

\section{Author's Role and Sources}

The author worked on the Logovista English--Japanese machine translation system from 1990 to 1994 and again from 2000 through 2012, and was the primary author of the core translation engine. The system was evaluated and marketed as a high-accuracy English--Japanese translator from the mid-1990s onward. This account is written in 2026, approximately thirteen years after the author’s last direct involvement; while it draws on surviving source code, version-control records, and documentation, some aspects necessarily reflect the author’s recollection.

\section{Business Context}

The Logovista English--Japanese machine translation project originated with Susumu Kuno, Professor of Linguistics at Harvard, who proposed building a commercial translation engine grounded in his linguistic work. In collaboration with Catena, a Japanese corporation, this led to the formation of Logovista on the Japanese side. Core software and linguistic development in the United States was carried out under contract by Language Engineering Corporation (LEC), based in Belmont, Massachusetts, where Kuno served as Chief Scientist.

Funding was secured in 1990, and LEC hired the author as its first employee that summer. Glenn Akers served as Chief Executive Officer of LEC. The first commercial release of the English--Japanese system occurred in Japan in 1992. Updated versions continued to be sold by Logovista for roughly two decades, through at least the end of 2012.

Around 2000, Logovista acquired LEC, after which U.S.-based development continued under the name Logovista US. Following the acquisition, development continued for an extended period within a smaller U.S.-based group focused on maintaining and incrementally extending the existing English--Japanese system.

\section{Architecture}

The system comprised a user-interface component and a core translation engine. The author’s work was confined to the engine, which accepted English text as input and produced Japanese text as output. The architecture of the engine was partially documented in U.S. Patent 5,528,491 (1996), ``Apparatus and method for automated natural language translation,'' inventors Susumu Kuno and Barton D. Wright.

The translation engine was based on explicitly authored grammar rules operating over a proprietary, hand-coded central dictionary. Additional dictionaries of proper names and technical terms for specialized domains were largely generated automatically from external sources. The core software was originally written in C and later ported to C++.

Most linguistic knowledge resided in external data files authored by linguists rather than in procedural code, although some linguistic decisions were inevitably embedded in the software. The principal linguistic resources were the grammar, multiple dictionaries, and a set of transformational rules referred to as the \textit{analyze} file. All such resources were stored as plain text and maintained under RCS version control, allowing incremental modification and rollback over time.

\subsection{English Analysis}

English input processing proceeded through three main stages:
\begin{enumerate}
\item A preparser segmented raw text into tokens corresponding roughly to words and punctuation.
\item A chart parser generated all syntactically valid parses licensed by the grammar.
\item Each syntactic parse was expanded into a richer representation incorporating semantic structure. Competing interpretations were evaluated using weighted contributions from multiple hand-coded scoring components (``experts''), and the highest-scoring analysis was selected for transfer. In practice, aggressive pruning heuristics were required to keep the number of candidate analyses computationally manageable.
\end{enumerate}

\subsection{Japanese Synthesis}

Japanese generation relied on information attached to the English analysis and grammar rules, together with additional transformational specifications:
\begin{enumerate}
\item Grammar rules specified how sister constituents should be reordered to produce appropriate basic word order in Japanese.
\item A separate rule file (\textit{analyze}), modeled on transformational grammar formalisms, applied further structural modifications.
\end{enumerate}

\section{Interpretation Scoring and Pruning}

The most technically challenging aspect of the system was selecting a single interpretation for translation from a directed acyclic graph representing all syntactically valid analyses of an input sentence. The number of distinct interpretations implicit in such graphs was often enormous; effective search spaces on the order of $10^{35}$ alternatives were not unusual.

Interpretation selection relied on scoring partial and complete analyses using multiple hand-coded evaluation components (``experts''), each contributing weighted preferences or penalties. Some of these were straightforward. For example, lexical preferences favored common word senses over rare ones, and grammar rules corresponding to frequent constructions were preferred over less common alternatives.

More informative constraints were semantic. The central dictionary encoded semantic restrictions and preferences on verb arguments, including subjects and various complements. Candidate analyses were evaluated by comparing these expectations against semantic properties associated with nouns, with penalties applied for mismatches and bonuses for especially well-matched combinations. In the presence of competing conjunction structures, semantic similarity among coordinated noun phrases played a significant role in ranking alternatives.

In practice, effective pruning was essential. Scoring was applied incrementally to partial analyses, allowing large regions of the search space to be discarded early. Without such pruning heuristics, exhaustive enumeration of interpretations would have been computationally infeasible.

\section{User-Guided Disambiguation}

Considerable effort was devoted to mechanisms for user-guided disambiguation, including interfaces for marking constituent structure and for selecting among alternative target-language realizations of individual words or phrases. Although these facilities were technically functional and exposed a large amount of internal structure, they were rarely used in practice by end users. In deployed settings, users overwhelmingly preferred fully automatic translation, even when the system’s output was imperfect.

\section{Development Procedure}

\subsection{Initial Coverage and Expansion}

Initial development focused on defining a comprehensive but finite set of English grammatical constructions and lexical possibilities, based on linguistic analysis rather than corpus statistics. The grammar reflected an attempt to enumerate the major structural patterns of English, while the central dictionary encoded a wide range of syntactic and semantic properties associated with individual lexical items.

As the system was extended to handle real-world text, this initial categorization proved insufficient. Coverage expansion repeatedly required the introduction of additional distinctions that had not been anticipated at the outset. In particular, verbs were classified into a large number of types based on complement structure and argument realization; on the order of forty distinct verb classes were ultimately distinguished, even when key prepositions or particles were treated as variables rather than fixed lexical items.

\subsection{Coverage Expansion and Regression Control}

Extending coverage was a continual objective, driven in part by customer feedback and by failures observed on real-world text. Addressing such failures typically required adding new grammar rules or expanding existing dictionary entries to account for previously unseen constructions.

Changes could not be made in isolation. Because correct translations emerged from competition among many ambiguous interpretations, improvements in one area frequently altered the relative ranking of analyses elsewhere. As a result, regressions—cases in which previously correct translations degraded—were common and expected.

To manage this, regression test sets totaling on the order of 10{,}000 English sentences with corresponding Japanese translations were accumulated over time. The system under development was repeatedly evaluated against this fixed test set to detect unintended changes in behavior. The most common tuning mechanism was adjustment of the weights associated with specific dictionary entries or scoring components, allowing relative preferences to be shifted without introducing new structural distinctions.

\textbf{Verification pending:} presence and completeness of regression test sets in the preserved archive.

\section{Limits from Ambiguity Growth}

Extending coverage by adding new lexical items was comparatively straightforward. In contrast, extending coverage by introducing new grammar rules or elaborating the entries for high-frequency words steadily increased structural ambiguity. Over time, this led to a growing incidence of regressions and reduced the flexibility with which individual problems could be corrected, as local changes increasingly had non-local effects on interpretation ranking.

\section{Preserved Artifacts}

When Logovista US ceased operations at the end of 2012, the author retained a copy of the core translation engine and associated development materials. These were originally provided to support potential technology transfer or licensing efforts, although no such transfer ultimately occurred.

The preserved materials include the complete core translation software (in C++), together with the principal linguistic resources, including the grammar, central dictionary, \textit{analyze} rules, and related files. In addition, the development system used internally as a driver for the core engine has been retained; this component was not part of the commercial product.

Version-control archives (RCS) are available for the software and linguistic resources, recording changes over time from at least the year 2000 onward, and possibly earlier. These archives make it possible to examine how coverage and behavior evolved in response to ongoing development and maintenance pressures.

The preserved materials are believed to include regression test sets used during development, consisting of English sentences paired with expected Japanese translations; confirmation of their completeness remains ongoing.

\section{Availability and Access}

The materials described here are preserved in private archival form by the author. At the time of writing, they are not publicly released. The archive was retained following the closure of Logovista US to support potential technology transfer or historical preservation, rather than as an active development base.

Access to the materials may be possible for qualified researchers under conditions consistent with intellectual property constraints and third-party rights. Any such access would necessarily exclude proprietary dictionaries or resources derived from external sources for which redistribution rights are unclear.

The intent of preserving these materials is to enable future historical or technical study of a large, long-lived, rule-based machine translation system, rather than to support continued commercial use or deployment.

\section*{References}

\begin{itemize}
\item S. Kuno and B. D. Wright. \textit{Apparatus and method for automated natural language translation}. U.S. Patent 5,528,491, 1996.
\end{itemize}

\end{document}